\documentclass[conference]{IEEEtran}
\ifCLASSINFOpdf
\else
\fi

\usepackage{times}
\usepackage{epsfig}
\usepackage{graphicx}
\usepackage{amsmath}
\usepackage{amssymb}
\usepackage[pagebackref=true,breaklinks=true,letterpaper=true,colorlinks,bookmarks=false]{hyperref}

\begin{document}
%
\title{Comparison of Semantic Segmentation Approaches for Horizon/Sky Line Detection}

\author{\IEEEauthorblockN{Touqeer Ahmad\IEEEauthorrefmark{1}, Pavel Campr\IEEEauthorrefmark{2}, Martin \v{C}ad\'{i}k\IEEEauthorrefmark{3}, George Bebis\IEEEauthorrefmark{1}}
\IEEEauthorblockA{\IEEEauthorrefmark{2}University of West Bohemia, Pilsen, Czech Republic}
campr@kky.zcu.cz
\IEEEauthorblockA{\IEEEauthorrefmark{1}Department of Computer Science and Engineering, University of Nevada, Reno, USA}
tahmad@nevada.unr.edu\\
bebis@cse.unr.edu
\IEEEauthorblockA{\IEEEauthorrefmark{3}  Brno University of Technology, Faculty of Information Technology, Centre of Excellence IT4Innovations, Czech Republic}
cadik@fit.vutbr.cz
}


\maketitle

\begin{abstract}
Horizon or skyline detection plays a vital role towards mountainous visual geo-localization, however most of the recently proposed visual geo-localization approaches rely on \textbf{user-in-the-loop} skyline detection methods. Detecting such a segmenting boundary fully autonomously would definitely be a step forward for these localization approaches. This paper provides a quantitative comparison of four such methods for autonomous horizon/sky line detection on an extensive data set. Specifically, we provide the comparison between four recently proposed segmentation methods; one explicitly targeting the problem of horizon detection\cite{Ahmad15}, second focused on visual geo-localization but relying on accurate detection of skyline \cite{Saurer16} and other two proposed for general semantic segmentation -- Fully Convolutional Networks (FCN) \cite{Long15} and SegNet\cite{Badrinarayanan15}. Each of the first two methods is trained on a common training set \cite{Baatz12} comprised of about 200 images while models for the third and fourth method are fine tuned for sky segmentation problem through transfer learning using the same data set. Each of the method is tested on an extensive test set (about 3K images) covering various challenging geographical, weather, illumination and seasonal conditions. We report average accuracy and average absolute pixel error for each of the presented formulation.
\end{abstract}


%
\IEEEpeerreviewmaketitle

\section{Introduction}
\label{sec:intro}

With the massive availability of geo-tagged imagery and increased computational power, geo-localization/geolocation has captured a lot of attention from researchers in computer vision and image retrieval communities. Significant progress has been made in urban environments with stable man-made structures and geo-referenced street imagery of frequently visited tourist attractions ~\cite{Hays08,Zheng09,Zamir10}. Recently some attempts have been made towards geo-localization of natural/mountain scenes which is more challenging due to changed vegetations, lighting and seasonal changes and lack of geo-tagged imagery. Typical approaches for mountain/natural geo-localization rely on mountain peaks and valley information, visible skylines, ridges or combinations of all three \cite{Baboud11,Baatz12,Tzeng13,Porzi14,Liu14,Chen15,Saurer16}. Sky/horizon line has been established to be a robust natural feature for mountainous images which can be matched with the synthetic skylines generated from publicly available terrain maps -- Digital Elevation Models (DEMs). Hence, the very first step in the geolocation pipeline for mountainous regions is to find the skyline in the given query image. However, most of the solutions for mountainous geo-localization rely on \textbf{user-in-the-loop} methods for skyline extraction where a user is required to mark/correct portion of the sky/horizon line \cite{Baatz12,Tzeng13,Chen15,Saurer16}. In addition to visual geo-localization and mountain image annotation/tagging, sky/horizon line has proven to be useful for various other applications e.g. UAV navigation \cite{Boroujeni12,McGee05,Thurrowgood09,Croon11,Ettinger02,Todorovic03}, vehicle navigation\cite{Ho14}, augmented reality \cite{Porzi14} and port security \cite{Fefilatyev06}. It should be noted that most of the earlier horizon/sky line detection approaches assume horizon to be a linear boundary; Hough transform was generally employed to find the line parameters subject to some cost function \cite{Ettinger02,McGee05,Todorovic03,Fefilatyev06,Liu14b}. Although linear horizon boundary could be of good use for UAV navigation, ship detection and/or port security; a non-linear sky segmentation is a must for geo-localization and hence the focus of this paper.\par 

\subsection{Related Work -- Mountainous Geo-Localization}

Using silhouette edge matching, Baboud et al. \cite{Baboud11} estimate the pose of camera relative to geometric terrain model (DEM) assuming known viewpoint and FOV estimates. Effectively a rotation $g \in SO(3)$ is searched which maps the camera frame to the terrain frame. They developed a robust silhouette matching metric to cope with inevitable noise affecting detected edges (compass edge detector is used). Since, a direct extensive search on $SO(3)$ based on their devised metric is quite expensive, that is why they also proposed a pre-processing search space reduction step based on spherical cross-correlation of 2D edge orientation vectors. They reported that 86\% of 28 images were correctly aligned belonging to two distinct mountain regions with matching error below $0.2^{\circ}$. Baatz et al. \cite{Baatz12} proposed a visual geo-localization pipeline based on bag-of-curvelets; where shape information is aggregated across the whole skyline of a query image and a similar configuration of shapes is searched in a large scale database of panoramic skylines (extracted offline from DEMs). In addition to encoded contourlets, the viewing direction for each descriptor is also saved which is used for on-the-fly geometric verification in an inverted file search framework. Since, they are comparing $10^{\circ} - 70^{\circ}$ views with $360^{\circ}$ panoramas, they redefine the weighted L1-norm to implement ``contains''-semantics instead of conventional ``equal''-semantics used for visual words (curvelets) matching. The most promising coarse estimate for the viewing direction (azimuth) is used to initialize ICP (keeping other two angles at zero) which determines full 3D rotation. The average alignment error between two visible horizons is used to re-rank the candidates in ICP framework. They reported an 88\% recognition rate on their challenging data set comprised of more than 200 images where determined position was within 1km radius of the ground truth. It should be noted that about half of their images required \textbf{manual interaction} at the sky segmentation stage.\par 
Somewhat similar to Baatz et al. pipeline \cite{Baatz12}, Tzeng et al. \cite{Tzeng13} proposed a localization approach for desert imagery. However, instead of curvelet features, concavity-based features across query and synthetic skylines are used for matching without any use of meta-data such as GPS, FOV and focal length. In contrast to \cite{Baatz12} where overlapping curvelet descriptors are generated for pre-defined angular width, they generate concavity descriptors around detected points of extreme curvature. Further a similarity transformation is applied on features to achieve scale and in-plane rotation invariance. The endpoint matching and feature’s shape matching is accomplished through geometric hashing and k-d trees respectively. Ranked database skylines from both matchers (endpoint and shape) are further refined using alignment error based on sampling of overlapping regions between query and database skylines. It should be noted that in their method, skyline in the query image is first roughly \textbf{marked by a user} and further refined by edge detection and dynamic programming framework as detailed in Lie et al. \cite{Lie05}.\par 
Porzi et al. \cite{Porzi14} also addressed the same image-to-world registration problem however in the context of an Augmented Reality based smart phone application. They first compute rough estimates for position and orientation from phone's on-board GPS and inertial sensors. These estimates are then refined by matching the skylines extracted from images taken by phone's camera and rendered from DEMs generated on a server. In principle their approach is closer to that of \cite{Baboud11} since they also assume roughly known position and orientation; however they rely on a learning based edge filtering approach which results in improved accuracy and computational cost desirable for a smart phone application. Based on the orientation estimates from device's inertial sensors, skyline detected from the phone's camera image and rendered profiles received from the server; \cite{Porzi14} define a search space around the rough estimate which is explored by Particle Swarm Optimization for refined orientation estimates. This is accomplished by maximizing the objective function based on the matching between the skyline contour and contours projected (pin-hole camera projection model) from profiles received from server.\par

\subsection{Paper Contribution}

The body of work on truly automatic non-linear horizon/sky line detection is rather limited and work comparing such methods is even rarer with the exception of Ahmad et al. who compare different formulation \cite{Ahmad14,Ahmad15a,Ahmad15b} of their approach against original approach of \cite{Lie05}. And quite recently the work by Porzi et al. \cite{Porzi16}; who proposed a small scale deep-learning architecture inspired from VGG  \cite{Simonyan14} for horizon line detection. They compared the performance of proposed network against \cite{Ahmad15} and \cite{Porzi14} etc. on CH1 data set \cite{Baatz12}. However, both of these comparisons are based on rather smaller data sets. To the best of our knowledge, this paper presents a first detailed quantitative comparison of truly automatic non-linear horizon/sky line boundary detection methods on a decent sized data set (about 3K images).\par
Another lacking aspect of sky segmentation literature being the use of accurate metrics to measure the accuracy of the detected boundary e.g. the question: ``on average how far is the segmented boundary from the ground truth boundary? '' is mostly not answered. This is true with the exception of few \cite{Ahmad15,Ahmad15a,Ahmad15b,Hung13,Porzi16} who have reported such measures. For example; Saurer et al.\cite{Saurer16} reported full automatic detection for 60\% of their images (total 948), however how good the detections were for these images was never mentioned. To address this issue we report both average pixel accuracy and mean absolute average distance between the found segmentations and the ground truths for each of the methods being compared.\par 

Specifically we provide a quantitative comparison for the following four methods: 
\begin{enumerate}
\item Ahmad et al. \cite{Ahmad15} proposed a horizon detection method inspired from \cite{Lie05} where instead of relying on edges, they used classification scores to provide confidence of horizon-ness. The given image is converted to a classification score image which is formulated as a multi-stage graph and a shortest path is found which conforms to a detected horizon boundary in the given image. 
\item For their visual geo-localization problem; Baatz et al. \cite{Baatz12} relied on a sky segmentation approach based on dynamic programming \cite{Lie05}, gradient magnitude and classification. They reported \textbf{human-involvement} for about half of the images in their original data set CH1 \cite{Baatz12}, the method is later refined and described in more details in their extended work \cite{Saurer16} where \textbf{human involvement} reduced to 40\% on CH2 data set. 
\item Long et al. \cite{Long15} are the first to propose training of Convolutional Neural Networks based on full scale images instead of conventionally used small-scale image patches. They proposed the ideas of deconvolution layers and fusion of finer and coarse levels to achieve semantic segmentation instead of instance/rectangular segmentation. Their proposed approach (FCN) has outperformed several popular deep-learning methods for segmentation and detection on various challenging data sets including PASCAL VOC, NYUDv2 and SIFTS Flow and has been widely used. 
\item SegNet \cite{Badrinarayanan15} also exploits the idea of deconvolutional layers and information fusion, however their decoder architecture is much denser compared to FCN\cite{Long15} and information is fused at several levels.  

\end{enumerate}

We adopt these deep-learning models for sky segmentation (horizon line detection) hence targeting a binary semantic segmentation problem. However, it should be noted that the non-sky class is more general i.e. the non-sky region could have many geographical, seasonal variations and similarly sky could have various illumination variations along with the clouds. Each of the first two method is trained on CH1 data set \cite{Baatz12} while models for the other two methods \cite{Long15,Badrinarayanan15} are adopted for sky segmentation problem and further fine-tuned. All four segmentation methods are compared on a single decent sized data set with significant geographical, seasonal, illumination and viewpoint variations. We report average accuracy and average absolute pixel errors as performance measures.\par

The remainder of the paper is organized as follows: The next section briefly describes each of the considered approaches for sky segmentation. Specific details regarding training or otherwise for each method are provided in section \ref{sec:specs}. Section \ref{sec:experiment} presents the training and test sets along with the performance metrics being used for evaluation. Results for each of the formulations and improvement due to further post-processings are listed in section \ref{sec:results} along with discussion. The paper is then concluded with insights and directions for the future work.


\section{Approaches Being Compared}
\label{sec:approaches}
For a standalone presentation, this section provides a brief overview of different segmentation approaches for horizon/sky line detection. Interested readers should consult actual papers for further details.  

\subsection{Ahmad et al. \cite{Ahmad15}}

Inspired from Lie et al. \cite{Lie05}; Ahmad et al. \cite{Ahmad15} proposed a dynamic programming based horizon line detection method where instead of relying on edge detection \cite{Lie05} and/or edge classification\cite{Ahmad13,Hung13,Porzi14}, a dense classification map is generated equal to the size of input image. This is a representative example of classical patch based training where a hand-engineered feature vector is fed to a classifier to predict the class probability for the central pixel of the patch. This essentially mimics semantic segmentation; however instead of being the class labels, the pixels in the output image reflect the probability of horizon-ness. Once this dense classification score image is computed, the problem is formulated as a graph search problem where a shortest path is searched from source node to sink node in M $\times$ N graph. Unlike, Lie et al.\cite{Lie05} who first perform edge detection and then define graph vertices based on edge pixels; the graph generated by Ahmad et al.\cite{Ahmad15} is dense and hence does not require any gap filling which is highly dependent on the tolerance-of-gap parameter as used by Lie et al.\cite{Lie05} and others\cite{Ahmad13,Hung13,Porzi14}. Instead, they \cite{Ahmad15} reduce the size of the graph by keeping a small number of minima for each column (graph stage).\par

Ahmad et al. used normalized pixel intensities as features and trained small scale SVM/CNN classifiers. The patch-based training is based on 9 images (about 6K instances of 16x16 patches) while the testing is conducted on about 120 images belonging to two different data sets, one targeted for rover localization and other based on images collected from the web. 

In other related approaches; the graph is formulated based on refined/classified edges which still requires gap-filling \cite{Ahmad13,Hung13,Porzi14}. Various classifiers and features have been investigated to reduce the number of edge pixels considerably so that a shallow multi-stage graph can be formed. The gradient magnitude and gradient difference to enforce smoothness along the horizon and combining classification scores, edge evidence and difference of gradient magnitudes etc. have also been explored. 

\subsection{Saurer et al. \cite{Saurer16}}
The approach of Saurer et al. is also based on dynamic programming however the energy function being minimized is more involved and tries to incorporate both data and smoothness constraints in a more adaptive manner.   They formulate the problem as foreground(non-sky)-background(sky) segmentation problem where a per-column highest foreground candidate is searched subject to minimization of data term and smoothness term in the energy function. The data term in one column evaluates the cost of all pixels below the candidate to be assigned to foreground class and all pixels above it to be assigned to the background class. The pixel-wise likelihoods are computed through the classifier trained on contextual and super pixel representation. The smoothness term is based on the assumption that all pixels on the contour should have a gradient orthogonal to the skyline. Their pipeline also allows the user to mark foreground/background strokes for challenging images where all the pixels above the marked stroke are assigned to background and those below the stroke to the foreground. 

Their training was based on 203 images from the CH1 set and testing was done using 948 images from CH2 set. They reported little to more user-involvement for about 40\% of the test images. Overall, the accuracy of their proposed pipeline increased 18\%  compared to one reported in earlier version of the paper \cite{Baatz12}.

\subsection{Long et al. \cite{Long15}}
Long et al. built fully convolutional neural network (FCN) that is able to semantically segment image into multiple classes. The network can take input image of arbitrary size and produce semantic labeling of corresponding size i.e. end-to-end training. This model exceeded other state-of-the-art methods for semantic segmentation. The authors adapted several structures of neural network models that were used for classification tasks (VGG net \cite{Simonyan14}, GoogLeNet\cite{Szegedy14}, AlexNet\cite{Krizhevsky12}) and fine-tuned them for the segmentation task. The networks were evaluated on PASCAL VOC, NYUDv2 and SIFT Flow datasets, where they achieved state-of-the-art results for multiclass segmentation. Specifically the conventional fully connected layers towards the right end of these networks are replaced with convolutional layers. The core of the FCN is the skip-layer architecture which combines the deep, coarse semantic information with shallow, fine appearance information. \par
\begin{figure}[h]
\begin{center}
\includegraphics[width=0.9\linewidth]{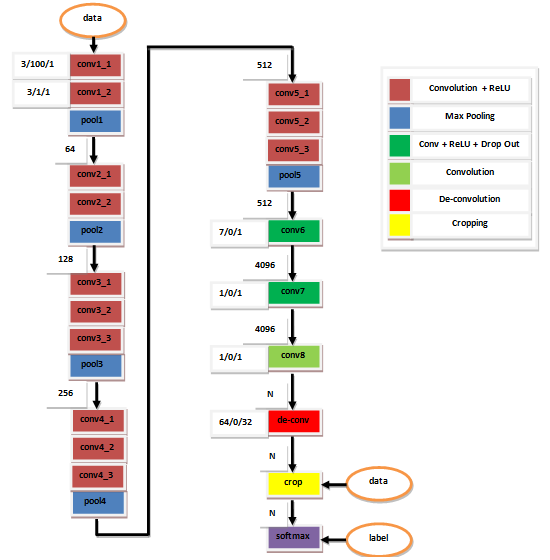}
\end{center}
\caption{VGG\cite{Simonyan14} network transformed into FCN32s\cite{Long15}}
\label{fig:long15}
\end{figure}

Figure \ref{fig:long15} shows the VGG network \cite{Simonyan14} transformed into FCN32s. Each of the convolutional (conv) layers is followed by an element-wise Rectified Linear Unit (ReLU) and a dropout layer (only in conv6 and conv7); color codings are provided to note the specific differences. For all the conv layers the receptive field, zero-padding and stride are of 3, 1 and 1 respectively except where explicitly noted as (F/P/S) to the left of the conv layer e.g. conv1\_1 has a padding of 100 and conv6 has a kernel of 7. Each module of the conv layers is followed by a max pooling layer with a filter of size 2 and stride 2. The number of output channels for each of the conv modules is noted with the number next to the arrow emerging from preceeding pooling layer. Conv layers conv6 and conv7 are each followed by a dropout layer with 50\% ratio i.e. 50\% of the neurons are dropped randomly in the respective layers to ensure generalization \cite{Srivastava14}. The conv8 layer is the compression layer responsible for compressing the 4096 channels to N channels where N is the number of classes. The de-convolution layer (de-conv) performs up-sampling while crop layer takes two inputs and crop the first according to the dimensions of the second. The softmax loss function is used to guide the stochastic gradient descent which takes the output of the crop layer and semantic label equivalent to the size of images.\par 
Given the input resolution of a conv layer the resolution of the output can be computed using Eq. \ref{eq:e1} while Eq. \ref{eq:e2} can be used for similar calculation for the de-conv layer; $I_r$ and $O_r$ are the resolutions of input and output to a layer while F, P and S indentify the  filter/kernel, zero-padding and stride respectively.  

\vspace{-2mm}

\begin{equation}\label{eq:e1}
O_r = \frac{(I_r-F+ 2P)}{S}+1 
\end{equation}

\begin{equation}\label{eq:e2}
O_r = S(I_r-1)+F-2P 
\end{equation}

Transfer learning \cite{Yosinski14} is an emerging trend in deep learning research where instead of training new network from scratch, existing networks are optimized and fine-tuned on one's own relevant data set. This is motivated due to the fact that training these networks is a time consuming task if done from scratch and realization that the earlier layers in any deep network are more general. In this paper we tested some of the FCNs for sky segmentation that were fine-tuned on different datasets by \cite{Long15}. We compare segmentation results using different approaches i.e. how to obtain sky segmentation from the multiclass output given by the networks into binary output (sky vs non-sky). Additionally, we have fine-tuned these existing networks on CH1 dataset to see if the existing multiclass segmentation network can be fine-tuned to perform better for the task of binary classification.\par

\subsection{Badrinarayanan et al. \cite{Badrinarayanan15}}

SegNet  is also motivated by the same principles as FCN and is based on fully convolutional layers and does not involve any fully connected layer, however it further focuses on maintaining sharp boundary delineation which is essential for pixel-wise segmentation of small/rare classes. Additionally, unlike FCN which requires the stage-wise training where a new decoder (fusion at multiple strides) is progressively added to the existing trained network (FCN-32s) and the resulting network (FCN-16s) is trained again; SegNet provides the capability of end-to-end training thanks to its decoder network. The encoder network in SegNet is exactly same as that of FCN i.e. 13 convolutional layers from VGG16 \cite{Simonyan14}, however it is followed by a decoder containing equal number of convolutional layers as in encoder and is the core of SegNet. Each decoder in the decoder network first upsample its input coming from the corresponding encoder (max-pooling indices) in the encoder network and then followed by learnable convolutional layers. The final convolutional layer is followed by a classification layer (softmax) same as in FCN. This decoder-decoder architecture of SegNet allows end-to-end training and crisp boundary delineation while FCN would have to rely on additional components/architectures e.g. CRFs and RNNs to achieve the similar objective \cite{Zheng15}.


\section{Specific Details}
\label{sec:specs}

\subsection{Edge-less Horizon Line Detection (DCSI)}
We replicated the approach of \cite{Ahmad15} and trained an SVM classifier based on CH1 data set. For each of the training images (203 total) the positive (horizon) key points are chosen along the ground truth horizon line while equal number of negative (non-horizon) key points are chosen from edge locations randomly which are not in a close vicinity of the ground truth horizon. A $16\times16$ normalized intensity patch is used as a feature vector. Unlike \cite{Ahmad15} the key points and features are extracted from the raw images instead of resizing them to a fixed resolution. The trained SVM classifier is used to generate a dense classification score map where a normalized score [0--1] for each pixel reflects its probability of horizon-ness. The dense image is taken as a dense graph and dynamic programming is applied to find a shortest path from left-most column (graph stage) to right-most column (graph stage) which conforms to a detected horizon boundary. It should be noted that Ahmad et al. \cite{Ahmad15} further make their graph sparse by keeping fixed number of minima for each stage; however for our implementation dynamic programming is run on a dense graph originally resulting from the classification score image.

\subsection{Automatic Labeling Environment (ALE)}
\emph{Automatic Labeling Environment~\cite{Ladicky10}} is an energy minimization-based 
semantic segmentation framework adopted for sky extraction by Saurer et al.~\cite{Saurer16}. 
Specifically, the energy is predicted by a pixel-wise classifier trained on contextual and superpixel feature 
representations. Multiple bag-of-words representations over the random set of 200 rectangles, and superpixels 
are used for contextual part, and superpixel part, respectively. The segmentation is obtained by 
minimizing the energy using dynamic programming (DP). 
With the personal advice of the authors, we implemented the algorithm~\cite{Saurer16}
into their Automatic Labeling Environment (ALE)~\cite{Ladicky10}. 
Similarly to the original paper~\cite{Saurer16} we set the number of bag-of-words clusters to 512 and
we train ALE using CH1 dataset~\cite{Saurer16}. 

\subsection{Fully Convolutional Neural Networks (FCNs)}
\label{subsection:fcns}
We used two different types of FCN models. PASCAL-context models were trained by the authors \cite{Long15} on object and scene labeling of PASCAL VOC, in three different resolution capabilities (FCN32s, FCN16s and FCN8s with the highest resolution). The models include both object and surface classes (59 classes, including class ``sky" and ``mountain"). This type of network predicts scores for each class at each pixel location. SIFT Flow models were trained for joint geometric (3 classes: ``sky", ``horizontal", ``vertical") and semantic (33 classes, including ``sky" and ``mountain") class segmentation and produces two separate scores. As all of the FCN models were trained for multiclass segmentation, we evaluated several methods to compute binary segmentations (sky vs non-sky) from scores being output by the networks. For PASCAL-context networks we compared scores for class 40 (mountain) and class 50 (sky) for each pixel. If the first score was higher the final class was set to ``non-sky", otherwise it was set to ``sky". For SIFT Flow networks there are more options, because the network provides two types of scores. We segmented a pixel as ``sky" in case the highest semantic score was for class 28 (``sky"), otherwise it was segmented as ``non-sky". Similarly for the geometric score, we segmented a pixel as ``sky" for case where the highest geometric score was for class 1 (``sky"), non-sky was segmented otherwise.

For the best performing models we fine-tuned the weights of the pretrained networks. The network structure and training parameters were kept the same as for the original networks. The CH1 dataset was used to provide input images and binary labels (sky vs non-sky). For the PASCAL-context pretrained network we modified binary labels so that the class indices correspond to correct classes in the network (class 40 denotes ``mountain", class 50 denotes ``sky"). For the SIFT Flow dataset, we had to convert source binary labels into two different target labels (geometric and semantic). Semantic label was set to class 28 (``sky") where the original label denoted sky, class 17 (``mountain") was used otherwise. Similarly, the geometric label was set to class 1 (``sky") where the original label denoted sky, class 3 (``vertical") was used otherwise. We expected such fine-tuned networks to perform better than the original pretrained networks. First reason is that the network is fine-trained with new unseen training data. The second is that the network is forced by the new input labels to predict only two classes (``mountain" and ``sky"), so that such a new network is specialized for the task of sky and non-sky segmentation.

\subsection{SegNet}

Unfortunately, the publicly available models for SegNet were trained on urban images, specifically for semantic road scene segmentation. We have investigated different available models, and adopted the best performing SegNet model (``driving webdemo") for our sky segmentation problem. As mentioned earlier, SegNet is trained with urban imagery, using it directly for mountainous sky segmentation does not make sense. So, instead of using SegNet model directly, we first fine-tuned it with CH1 data set. Another disadvantage of segnet is that both input image and output segmentation have fixed resolution (480x360), so we have to resize the input to this size for training and later resized the 480x360 segmentations to the original sizes of respective images.

\begin{figure}
\begin{center}
\includegraphics[width=0.9\linewidth]{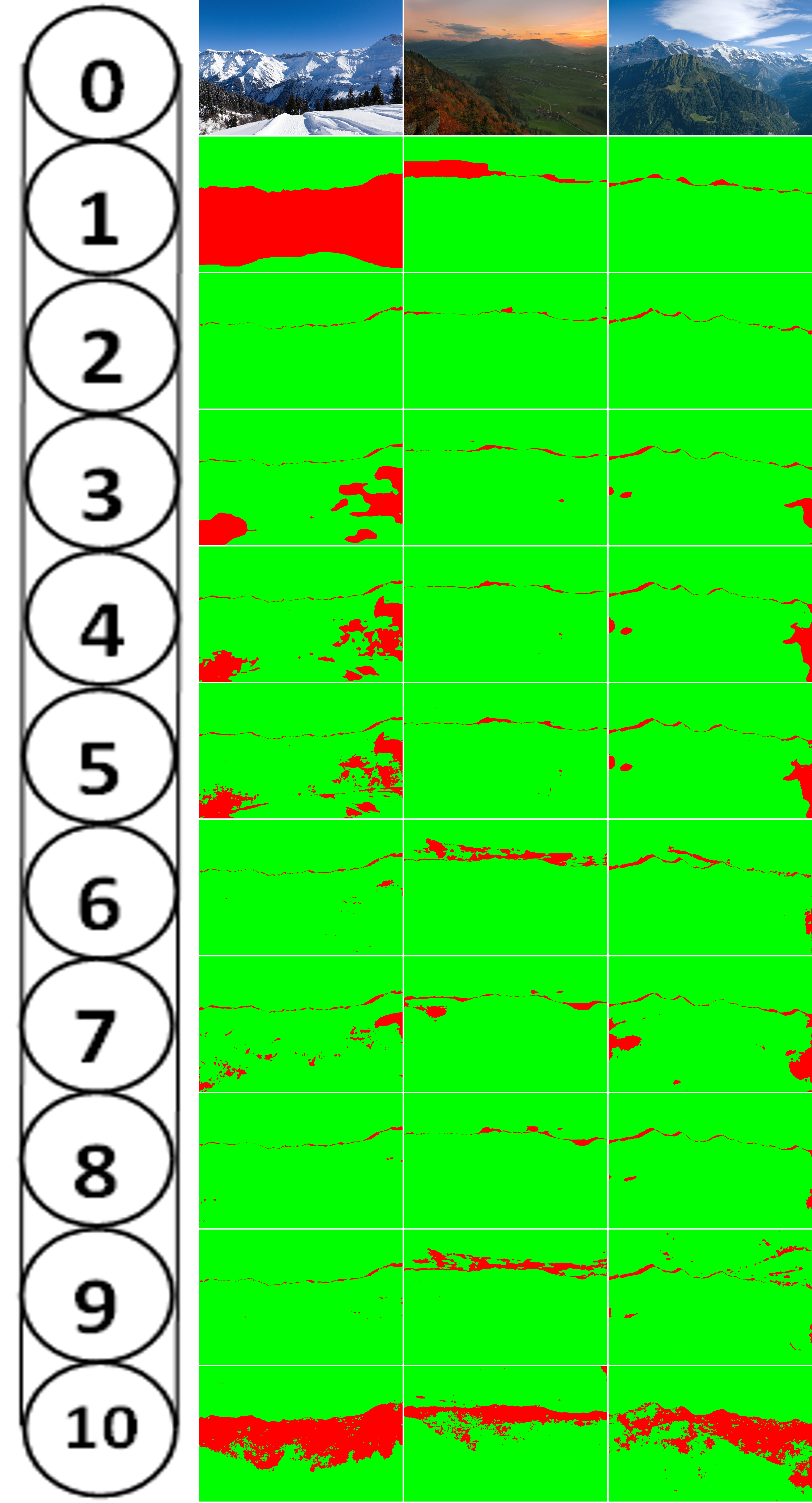}
\end{center}
\caption{Some visual results for segmentation, green -- correctly classified, red -- miss-classified: (0) sample images from our test set, (1) DCSI, (2) ALE, (3) FCN32s-Pascal, (4) FCN16s-Pascal, (5) FCN8s-Pascal, (6) FCN8s-Pascal-CH1, (7) FCN8s-SiftFlow-geometric (8) FCN8s-SiftFlow-semantic, (9) FCN8s-SiftFlow-semantic-CH1 and (10) SegNet-CH1.}
\label{fig:samples}
\end{figure}


\section{Experimental Details}
\label{sec:experiment}

\subsection{Data Sets}
Our test set is comprised of 2895 mountainous images~\cite{brejcha17geopose3k} which have been acquired from Flickr and is a subset of an even bigger data set which has been made publicly availabe previously \cite{Cadik15}. The GPS locations and camera intrinsic are used to access the relevant Digital Elevation Maps (DEMs) which are then rendered using a conventional OpenGL utility to develop the ground truth segmentations. These ground truth segmentations are used to compute the error metrics defined next. For training DCSI\cite{Ahmad15}, ALE\cite{Saurer16} and fine-tuning FCNs\cite{Long15} and SegNet\cite{Badrinarayanan15}, we use the publicly available CH1 data set \cite{Baatz12}. It should be noted that both training and test sets contain images with various resolutions. The deep-learning platform Caffe has been extensively used for training the deep architectures -- FCN and SegNet.

\subsection{Accuracy}
We use pixel-wise mean accuracy to establish the performance of the approaches being compared for sky segmentation. Specifically, for each of the images we compute what fractions of pixels have been correctly classified by an approach. The equation below shows the computation of mean accuracy for the test set: 
\begin{equation}
DC = \frac{1}{N_{set}}\sum_{i=1}^{N_{set}}\frac{{N_c}^i}{{N_t}^i}, 
\end{equation}   
where $N_{set}$ is the total number of images in the data set, ${N_c}^i$ and ${N_t}^i$ are the number of correctly classified pixels and number of total pixels in the image i. For an ideal 100\%  classification accuracy this measure should result in a perfect 1. 

\subsection{Average Absolute Pixel Distance}

The second measure being computed is the average absolute pixel distance between the ground truth and resultant segmentation. This has previously been suggested by \cite{Ahmad15} and later adopted by others \cite{Porzi16}. If a method generates an accurate solution the skyline should be consistent with the ground truth i.e. the vertical distance of the pixels belonging to the two boundaries should be minimized. This is measured through absolute vertical pixel-wise distance. Mathematically:  
\begin{equation}
S=\frac{1}{N_{set}}\sum_{i=1}^{N_{set}}\Bigg(   \frac{1}{N}\sum_{j=1}^{N}{|{P_{d(j)}}^i-{P_{g(j)}}^i|}\Bigg), 
\end{equation}
where $P_{d(j)}$ and $P_{g(j)}$ are the positions (rows) of the detected and true horizon pixels in column $j$ and $N$ is the number of columns in the test image. One drawback of this measure is the underlying assumption of having exactly one pixel at each stage (column) which is not necessarily true for steep peaks. However, since same measure is computed for all the methods; the effect of this assumption should be averaged out. A perfect alignment of detected skyline with the ground truth skyline should result into a zero for this error measure. 


\section{Results}
\label{sec:results}

In the first set of experiments, we report the results for all the considered formulations without any post-processing. For DCSI and ALE; the classifiers are trained on CH1 data sets while different flavors of FCN models are either adopted as-it-is for sky segmentation or have been further trained through transfer learning on CH1 data set as noted in section \ref{subsection:fcns}. Table \ref{tab:table1} lists both considered metrics for each of the formulations. The number after FCN i.e. (8s,16s or 32s) identifies the resolution at which the coarse semantic information has been fused with fine appearance information in the FCN architecture, SiftFlow or Pascal key-word identifies which data set has been used to train the model and an additional CH1 follows if the model has been fine-tuned further on CH1 data set. In case of SiftFlow, it is further distinguished between geometric (g) and semantic (s) models. Figure \ref{fig:samples} shows some visual results.

\begin{table}
\centering \caption{Performance of different formulations on our test set (2895 images).}
\begin{tabular}{|c|c|c|c|}
\hline
\textbf{Approach} & {\textbf{Accuracy}} & \multicolumn{2}{c|}{\textbf{Pixel Distance}}\\  \cline{3-4}
 &  & $\mu$  & $\sigma$  \\  \cline{1-4}
{\textbf{FCN8s-Pascal}} 								&	$0.9083$	&	$29.886$	&	$50.721$\\ \hline 
{\textbf{FCN16s-Pascal}} 							&	$0.9071$	&	$30.187$	&	$50.616$\\ \hline 
{\textbf{FCN32s-Pascal}} 							&	$0.9015$	&	$31.160$	&	$50.714$\\ \hline 
{\textbf{FCN8s-SiftFlow-g}} 							&	$0.9266$	&	$37.028$	&	$56.481$\\ \hline 
{\textbf{FCN8s-SiftFlow-s}} 							&	$0.9438$	&	$37.937$	&	$67.608$\\ \hline 
{\textbf{Horizon-ALE-CH1}} 							&	$0.9428$	&	$44.669$	&	$87.430$\\ \hline 
{\textbf{Horizon-DCSI-CH1}} 							&	$0.8756$	&	$99.425$	&	$160.516$\\ \hline
{\textbf{SegNet-CH1}} 								&	$0.8290$	&	$90.385$	&	$81.528$\\ \hline  
{\textbf{FCN8s-SiftFlow-s-CH1}} 						&	$0.9379$	&	$61.502$	&	$96.006$\\ \hline  
{\textbf{FCN8s-Pascal-CH1}}							&	$0.9285$	&	$68.283$	&	$97.626$\\ [1 ex]
\hline
\end{tabular}
\label{tab:table1}
\end{table}

\subsection{Post-Processing}

Post-processing of the binary segmentation images can further improve the segmentation quality. As seen from examples in figure \ref{fig:samples}, some methods are able to find horizon accurately, but resulting segmentation images are not physically possible in reality. One example is to have large sky area surrounded by non-sky area (``hole in an object"), which is rare to happen in physical world. Another example is to have non-sky area surrounded by sky (``flying object"). Several post-processing methods that reflect physical world properties can be designed to improve quality of the resulting segmentation images. We adopted the following two simple post-processing approaches to further enhance the segmentation results.\par 
It should be noted that not all the methods would benefit from such prost processing; specifically DCSI and ALE -- as these methods employ Dynamic Programming to find crisp boundaries between sky and non-sky regions and mostly do not suffer due to miss-classification holes. Nonetheless, all the segmentations have been post-processed for consistency.

\subsection{Post-processing I}

The first post-processing method uses two basic binary image processing operations. The first operation fills all holes, sky areas that are fully surrounded by non-sky areas are replaced by non-sky area. The second operation removes small non-sky objects, specifically all small non-sky objects that have area below 50\% of the largest non-sky object are removed, i.e. are replaced by sky label. The improvements due to this post-processing are listed in table \ref{tab:table2}. 

\begin{table}
\centering \caption{Segmentation improvment due to Post-processing I}
\begin{tabular}{|c|c|c|c|}
\hline
\textbf{Approach} & {\textbf{Accuracy}} & \multicolumn{2}{c|}{\textbf{Pixel Distance}}\\  \cline{3-4}
 &  & $\mu$  & $\sigma$  \\  \cline{1-4}
{\textbf{FCN8s-Pascal}} 								&	$0.9108$	&	$32.161$	&	$57.510$\\ \hline 
{\textbf{FCN16s-Pascal}} 							&	$0.9086$	&	$32.888$	&	$58.193$\\ \hline 
{\textbf{FCN32s-Pascal}} 							&	$0.9011$	&	$33.534$	&	$57.588$\\ \hline 
{\textbf{FCN8s-SiftFlow-g}} 							&	$0.9296$	&	$34.975$	&	$53.334$\\ \hline 
{\textbf{FCN8s-SiftFlow-s}} 							&	$0.9446$	&	$31.399$	&	$55.052$\\ \hline 
{\textbf{Horizon-ALE-CH1}} 							&	$0.9403$	&	$43.959$	&	$86.038 $\\ \hline 
{\textbf{Horizon-DCSI-CH1}} 							&	$0.8727$	&	$99.742$	&	$160.252$\\ \hline  
{\textbf{SegNet-CH1}} 								&	$0.8279$	&	$114.893$	&	$99.021$\\ \hline  
{\textbf{FCN8s-SiftFlow-s-CH1}} 						&	$0.9421$	&	$37.947$	&	$69.435$\\ \hline  
{\textbf{FCN8s-Pascal-CH1}}							&	$0.9351$	&	$41.596$	&	$71.707$\\ [1 ex]
\hline
\end{tabular}
\label{tab:table2}
\end{table}

\subsection{Post-processing II}
Similarly, two operations are used for the second post-processing method. The first operation removes small non-sky objects in the same way as in the first method. The second operation is column horizon detection, which finds first non-sky pixel label in a column from the top and sets all pixels below as non-sky pixel, i.e. first non-sky pixel in a column from the top defines the horizon. Table \ref{tab:table3} shows the improvements due to this post-processing approach.

\begin{table}
\centering \caption{Segmentation improvment due to Post-processing II}
\begin{tabular}{|c|c|c|c|}
\hline
\textbf{Approach} & {\textbf{Accuracy}} & \multicolumn{2}{c|}{\textbf{Pixel Distance}}\\  \cline{3-4}
 &  & $\mu$  & $\sigma$  \\  \cline{1-4}
{\textbf{FCN8s-Pascal}} 								&	$0.9551$	&	$32.161$	&	$57.510$\\ \hline 
{\textbf{FCN16s-Pascal}} 							&	$0.9539$	&	$32.888$	&	$58.193$\\ \hline 
{\textbf{FCN32s-Pascal}} 							&	$0.9520$	&	$33.534$	&	$57.588$\\ \hline 
{\textbf{FCN8s-SiftFlow-g}} 							&	$0.9491$	&	$34.975$	&	$53.334$\\ \hline 
{\textbf{FCN8s-SiftFlow-s}} 							&	$0.9563$	&	$31.399$	&	$55.052$\\ \hline 
{\textbf{Horizon-ALE-CH1}} 							&	$0.9411$	&	$43.959$	&	$86.038$\\ \hline 
{\textbf{Horizon-DCSI-CH1}} 							&	$0.8743 $	&	$99.742$	&	$160.252$\\ \hline  
{\textbf{SegNet-CH1}} 								&	$0.8437$	&	$114.893$	&	$99.021$\\ \hline  
{\textbf{FCN8s-SiftFlow-s-CH1}} 						&	$0.9486$	&	$37.947$	&	$69.435$\\ \hline  
{\textbf{FCN8s-Pascal-CH1}}							&	$0.9432$	&	$41.596$	&	$71.707$\\ [1 ex]
\hline
\end{tabular}
\label{tab:table3}
\end{table}



\subsection{Discussion}
While looking at the results reported in table \ref{tab:table1} and considering only the accuracy first, FCNs trained on SiftFlow clearly outperform the rest of the formulations. Surprisingly, the second best is the ALE approach which is a non-deep-learning method. This could be due to the fact that ALE is a well crafted approach for the problem of sky segmentation and may not generalize well to other segmentation applications unlike FCNs. The FCN8s -- with the finest resolution outperforms the others with coarser fusion (FCN16s, FCN32) which follows the results reported by authors \cite{Long15}. Interestingly, SegNet has performed very poorly, even worse than the simple SVM based DCSI method. This is contradictory to the results reported by authors \cite{Badrinarayanan15} for semantic road scene segmentation. It must be due to the fact that SegNet models made publicly available have been trained on urban street view data sets unlike FCNs. Another drawback of SegNet being the need to have images of fixed resolution which is not the case in FCN.\par
At first, the average absolute pixel error is not very consistent with the accuracy measure. Based on this measure alone, the poorly performing methods (i.e. DCSI and SegNet) can be readily identified and follow the observations made based on accuracy alone. However, from table \ref{tab:table1} it follows that FCN8s-Pascal performs better than FCN8s-SiftFlow which is not the case considering accuracy. This interesting contradiction clears out in table \ref{tab:table3} when FCN models based on Pascal benefit from post-processing II and both FCNs have very close average pixel errors.\par
Post-processing has proven to be helpful for all the FCN formulations while it does not have much effect on dynamic programming based methods -- DCSI and ALE which is inevitable. Although both post-processing methods impacted the segmentation results positively, the improvement due to second method of post-processing has been overall more effective.


\section{Conclusion}
We have provided a quantitative comparison of four different segmentation methods for mountainous sky segmentation: two of these methods belong to classical feature learning and patch-wise classifier training category while other two are instances of deep-learning networks -- recently proposed for semantic segmentation. We train the classifiers for first two methods using a publicly available data set, while deep-learning architectures are fine-tuned using the same data set through transfer learning. The segmentation results are further post-processed to improve the segmentation. The formulations are compared using mean classification accuracy and average absolute distance of the segmented boundary from true horizon. It should be mentioned that this is a first quantitative comparison of autonomous non-linear skyline detectors on an extensive data set.\par 
The Fully Convolutional Network (FCN) has proven to be best performing method for sky segmentation of mountainous imagery with ALE being the close second best. An obvious future direction would be investigating these segmentation methods and the data set for mountainous visual geo-localization.

\section*{Acknowledgements}
{ \footnotesize
This research was supported by Ministry of Education, Youth and Sports of the Czech Republic, project No. LO1506 and from the National Programme of Sustainability (NPU II); project IT4Innovations excellence in science - LQ1602. The work was also supported by the Technology Agency of the Czech Republic by project TE01020415 ``V3C" and by SoMoPro II grant (financial contribution from the EU 7 FP People Programme Marie Curie Actions, REA 291782, and from the South Moravian Region). %
At UNR, this work is supported by NASA EPSCoR under cooperative agreement No. NNX11AM09A and, in part by NSF PFI.
The content of this article does not reflect the official opinion of the European Union. Responsibility for the information and views expressed therein lies entirely with the authors.
}

\end{document}